\documentclass[journal]{IEEEtran}

\usepackage{cite}
\usepackage[T1]{fontenc}
\usepackage{booktabs}
\usepackage{amsmath,amsfonts}
\usepackage{multirow}

\usepackage[center]{caption}
\ifCLASSINFOpdf
  \usepackage[pdftex]{graphicx}
\else
\fi

\begin{document}
\bstctlcite{IEEEexample:BSTcontrol}

\title{Siamese based Neural Network for Offline Writer Identification on word level data}
\author{Vineet~Kumar and ~Suresh~Sundaram
\thanks{The authors are associated with Department of Electronics and Electrical Engineering, Indian Institute of Technology Guwahati, Guwahati 781039, India
 (email: vineet18@iitg.ac.in, sureshsundaram@iitg.ac.in).}}% <-this % stops a space

\maketitle

\begin{abstract}
Handwriting recognition is one of the desirable attributes of document comprehension and analysis. It is concerned with the document writing style and characteristics that distinguishes the authors. The diversity of text images, notably in images with varying handwriting, makes the process of learning good features difficult in cases where little data is available. In this paper we propose a novel scheme to identify writer of a document based on input word image. Our method is text-independent and does not impose any constrain on the size of the input image under examination. To begin with,  we detect crucial components in handwriting and extract regions surrounding them using Scale Invariant Feature Transform (SIFT). These patches are designed to capture individual writing features (comprising of allograph, character or combination of characters) that are likely to be unique for an individual writer. These features are then passes through a deep Convolutional Neural Network (CNN) in which the weights are learned by applying the concept of Similarity learning using Siamese network. Siamese network enhances the discrimination power of CNN by mapping similarity between different pairs of input image. Features learned at different scales of the extracted SIFT keypoints are encoded using Sparse PCA, each components of the Sparse PCA is assigned a saliency score signifying it's level of significance in discriminating different writers effectively. Finally the weighted Sparse PCA corresponding to each SIFT keypoints is combined to arrive at a final classification score for each writer. The proposed algorithm evaluated on  two publicly available databases (namely IAM and CVL) and is able to achieve promising result, when compared with other deep learning based algorithm.    

\end{abstract}

\IEEEpeerreviewmaketitle
\section{Introduction}
 \IEEEPARstart{H}{umans} have used writing to express themselves from the dawn of time. It is a distinguishing characteristic that allows a person to be identified. In other words, handwriting is a beahvioural biometric \cite{Chi:2018} feature that is unique in nature and is comparable to fingerprints in terms of relevance. Handwritten document analysis offers a lot of promise in  various applications such as forensic analysis \cite{Fernandez:2010}, historical document analysis \cite{Bulacu:2007}, \cite{He:2014} and security \cite{Faundez:2020}. \par

Writer Identification is associated with finding the authorship of an unknown document from a set of reference documents stashed in the database by providing a list of probable writers. Prior to the advance of modern technology handwriting analysis was mostly done manually by expert professionals, requiring domain knowledge to analyze and select the discriminating set of patterns. Such, analysis sometimes may not be conclusive owing to the difference of interpretation by various experts. Also, in a situation where a large amount of handwritten data needs to be examined manual examination  becomes a tedious process. An automated writer identification system helps us in mitigating some of these issues by extracting writer features directly from the raw data, thus eliminating the need of extensive manual examination of the document resulting in  making the whole process fast and efficient. Over the years a variety of options have been explored to capture writer-specific features effectively. Based on the literature survey these writer features can be grouped into three categories: texture based, shape based, and deep learning based methods.\par   

 Texture-based approach interprets handwriting as a sequence of texture, these texture information can be extracted either in the form of frequency, or as a spatial feature distribution. Methodologies relying on frequency domain consider handwriting as a combination of textures and employ various frequency domain operations such as Gabor filters \cite{SAID:2000,Helli:2010}, wavelet transformation \cite{Zhenyu:2008,Shen:2002} to characterize a writer information.\par
 
Opposed to frequency based approach to characterize the texture information, spatial features based technique views handwriting as a collection of edges and contour employing statistical technique on these  structures to represent a writer features. Some of the most commonly used spatial distribution are gray level co-occurrence matrices \cite{Djeddi:2010}, edge direction and edge hinge distribution \cite{Bulacu:2003}, oriented basic image features (oBIF) \cite{Newell:2014}, local phase quantization (LPQ) and local binary (ternary) patterns (LBP/LTP) \cite{Bertolini:2013,Hannad:2016}, scale invariant feature transform (SIFT) \cite{Wu:2014,Khan:2018}. \par  

In shape-based methods a handwritten sample is split into small segments based on specific features such as edges, corners, junctions etc. \cite{Akram:2019,Schomaker:2015}. These, segmented characters are clustered to form a code-book \cite{Bensefia:2002,Ghiasi:2013,Jain:2011,Schomaker:2004,Siddiqi:2010}, which are subsequently used to project the handwriting sample to describe the writer feature. \par

Recent years have seen a substantial change in approach towards image classification and computer vision related problem. The traditional hand-crafted feature descriptor are now being phased out in favour of deep learning based system. These deep learning-based feature representation approaches often outperform hand-designed feature descriptors in terms of recognition performance as they can learn data dependent features automatically from the training as opposed to hand-crafted features which require domain knowledge to extract set of discriminating writer features. The concept of deep learning in the field of writer identification was introduced by Fiel \textit{et.al} \cite{Fiel:2015}. In their work they trained a convolution network consisting of five convolution and three fully connected layers on segmented text image. The features extracted form the penultimate layer of the trained CNN network is used to construct a feature vector following which   ${\chi}^2$ distance is used to get identity of the writer. Following this work a number of similar studies \cite{Christlein:2015,Xing:2016,Christlein:2017} explored the effectiveness of features learned by ConvNets for the purpose of writer identification. For increasing the effectiveness of Convolution network in learning robust features to be applied in the field of writer identification  many modification were carried out. Christlein \textit{et.al} \cite{christlein:2017(b)} used surrogate class generated by clustering the training dataset to train the convolution network in an unsupervised manner. Chen \textit{et.al} \cite{chen:2019} used Semi-supervised method to train the network by incorporating extra unlabelled data along with the original labelled data. Yang \textit{et.al} \cite{Yang:2016} taking inspiration from the concept of dropout introduced the concept of Drop Stroke, thus modifying the original image by randomly removing some of the strokes form it. These combined images consisting of  modified and original image is used to train the network. Some of the recent work \cite{Sulaiman:2019,Semma:2022,semma:2021} focuses on extracting features generated form the intermediate layers of a convolution layer to encode the writer descriptor.\par

\begin{figure*}[ht]
 \centering
\includegraphics[width=.5\textwidth]{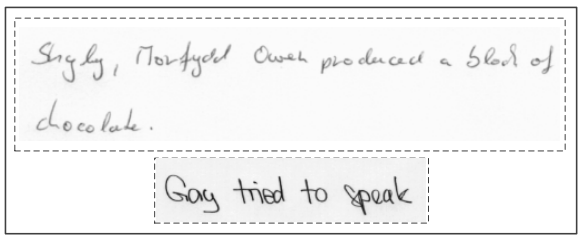}
\caption{Example of Handwritten samples having small amount of text data \cite{Wu:2014}}
 \label{fig:limted word}
\end{figure*}

Most of the preceding work discussed till now uses text-line/page-level data to extract features using deep-leaning/ handcrafted based approach to incorporate handwriting style related information between different individuals. However, in many cases the access to handwritten data  is limited (containing a few words) as depicted in figure (\ref{fig:limted word}) based on which decision needs to be made. In such a scenario writer identification techniques discussed above may find it difficult to discriminate between different writers owing to the availability of limited writer-related style information. In order to address this issue   He \textit{et.al} \cite{He:2018} proposed a multitasking framework consisting of two parallel stream of two CNN network applied of single word image. This framework enhances the writer information by assimilating the features learned from an auxiliary task into the main task over the shared feature representation. In a subsequent work \cite{He:2020} proposed a network called FragNet, which uses the concept of Feature pyramid \cite{Lin:2017} to extract fragments of variable size around a  feature map in a convolutional layer. These fragments are processed simultaneously using a separate neural network based on which writer identity is established. In a later extension of their work \cite{He:2021} they exploited the spacial relationship between the fragments using a  recurrent neural network (RNN). 

\section{RESEARCH FRAMEWORK}
In this paper, we propose a Dissimilarity based model for extracting writer features and identifying the authorship of handwritten document based on word image. The concept of Dissimilarity learning is inspired by one of the major behavioural trait of human learning involving the idea of similarity or semblance \cite{tversky:1977}. It is majorly used in problem involving multiclass classification \cite{Khan:2018} and in applications involving samples which tend to have discernible patterns such as  Handwritten text \cite{duin:2009}. The concept of dissimilarity involves estimating unknown samples identity based on it's similarity from the known sample. \par

In the present work we use a variant of neural network called Siamese network, which instead of classifying an input into fixed set of outputs (as done in most of the deep learning based classifiers) learns embedding, which aids in determining how similar or distinct different sets of objects are to one another. These type of network require less data for training thus making it a popular choice in many diverse applications, such as face recognition \cite{Roy:2021,Zhang:2018}, signature verification \cite{Sounak:2017} etc. \par

Contrary to the widely practiced methodology of feeding raw image or randomly segment patches of handwritten image to a neural network, we employ a key point detection algorithm to extract important keypoint regions in an image.  Since, discriminatory feature of a handwritten character is not unique and varies within across a language \cite{Tan:2010}, therefore in contrast to random sampling regions around these detected keypoints could be more effective  in feature learning. In the proposed work, we have utilized SIFT \cite{Lowe:2004} algorithm to locate various regions in an image containing informative writing patterns using multi-scale analysis. These extracted image patches are subsequently fed a CNN network trained on Omniglott datset \cite{Lake:2015}. \par

Corresponding to each fragment, a fixed sized embedding is generated. These fixed sized embedding is passes through an encoding block which modifies them into a sparse representation thus increasing the separability of the input samples. In order to further enhance the separability  we assign a weight to each component of the sparse representation, which helps in quantifying the degree of importance assigned to each component of the sparse representation. The methodology employed to assign a significance level to each sparse component is discussed in the subsequent sections.\par

The modified sparse representation corresponding to each writer word fragment is used to train an SVM classifier. All the SIFT fragments corresponding to a word image when passed though a trained SVM classifier generates an output score, which is then subsequently accumulated to establish the identity of a writer.\par

\subsection{Block schematic of Our proposal}
\begin{figure*}[t] 
 \centering
\includegraphics[width=\textwidth]{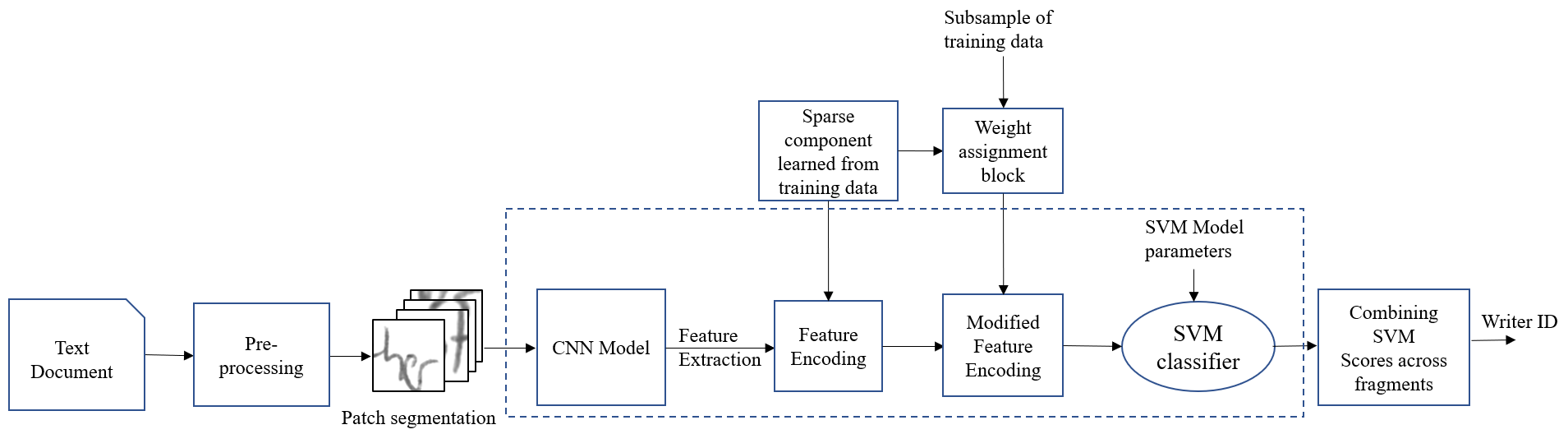}
\caption{Pictorial overveiw of the key steps in our proposed system.}
 \label{fig:outline}
\end{figure*}

 Fig.\ref{fig:outline} presents the overview of our proposed algorithm. The input text document is passed through a preprocessing module. This preprocessed output is then fed to a patch segmentation module resulting in extraction of important regions from the input image, which are subsequently fed to a trained CNN network for the purpose of feature representation. The sparse based technique, whose component is trained using the subset of training data connected to the writer identification system, is used to encode the output feature vector acquired from the CNN block. Each component of the sparse vector is subsequently assigned a significance score based on relative entropy based approach. The updated descriptor generated for each image fragment is send to a trained SVM classifier, which generates a set of classification score relative to each enrolled writer. These set of operations enclosed in the dotted block is performed for each fragment extracted form the patch segmentation module, following which the SVM scores generated across the fragments are accumulated for establishing the identity of the writer. \par

 The rest of the paper is organized as follows. In section \ref{sec3}, we discuss the pre-processing step consisting of background noise removal along with word segmentation. In section \ref{sec4} we describe the process of fragment generation form segmented word image using SIFT algorithm. This is followed by a detailed discussion about the Siamese based convolution network used for feature extraction in section \ref{sec5}. In section \ref{sec6}, we explain the concept of sparse based PCA encoding. Subsequently the methodology employed for assigning a significance score to individual sparse component of the PCA is discussed in section \ref{sec7}. In section \ref{sec8} the modified writer descriptor generated by incorporating the significance score is elaborated. The Datset used to evaluate our work along with the training and testing strategy employed is discussed in section \ref{sec9}. Section \ref{sec10} provides a set of experimental analysis related to our proposed work. Finally we conclude our paper in section \ref{sec11}.

\section{PREPROCESSING}\label{sec3}
The input text image is first converted into gray-scale following which the below mentioned operations are performed.
\begin{itemize}
    \item \textbf{Noise removal}: During the conversion of handwritten document into digital format an image may be degraded. The common cause of degradation in the quality of handwritten scanned image include noise encountered due to quality of the paper, ink blot and fading, document aging, extraneous marks, noise from scanning, etc. In order to remove such artifacts we employ a Gaussian filter based threshold operation. Based on the threshold value pixel is either retained or discarded. 
    \item \textbf{Word segmentation}: Following the Noise removal process, word regions are extracted form the input handwritten document. This step forms an important part of our algorithm as the overall performance of the system depends on this. Majority of the writer identification algorithm rely on extracting handwriting contours or allograph fragments form page/text-line level input data, these features are unable to represent structural similarity across allograph of the same word. In studies carried out by Wu.\textit{et.al} \cite{Wu:2014}, it has been observed that features extracted from word level data have a strong ability to learn writer level features effectively, as they form a stable structure. Based on their observation we extract extract word regions from an input image using an isotropic LoG filter based approach as in discussed in \cite{Wu:2014}. 
\end{itemize}
Following the preprocessing operation, a handwritten document is divided into many word regions, which are used for feature extraction in the subsequent stages. 
\section{PATCH EXTRACTION USING SIFT ALGORITHM}\label{sec4}
 Most of the deep learning algorithm use dense sampling (with or without overlap) \cite{Rehman:2019,Javidi:2020} in order to extract handwriting blocks from a given sample. However, based on the works carried out by Tan \textit{.et.al} \cite{Tan:2010}, it can be inferred that not all segmented regions (and their combinations) are equally discriminative for characterizing a writer. Thus, rather than extracting random patches from a sample, extracting regions around a specific interest point could be more effective in characterizing a writer feature. The purpose of interest point detection is to offer detailed spatial information about an image's structure in a way that is repeatable, mathematically sound, and stable even when the picture is perturbed. Interest point detection has found a great application in area of writer identification \cite{Newell:2014,Sharma:2015,Wu:2014,Schomaker:2015}. In the present work we have used SIFT algorithm to locate and extract interest point form the input image. \par
 
Since, its conceptualization SIFT \cite{Lowe:2004} has found a wide variety of application in many computer vision applications such as object detection, object localization, image stitching. An important property that differentiates it from other keypoint detector algorithm (such as Harris corner detector) is its ability to detect important features which are invariant to  image translation, scaling, and rotation. SIFT algorithm uses Difference of Gaussian (DoG) pyramid to search key points at different scales. Among a large number of detected keypoints, stable keypoints are extracted by selecting local maximum spots in the image pyramid's 3D neighbourhood. Following key-point localization the prevailing orientation for each key point is identified to achieve the rotation invariance. The extracted keypoint along with its orientation is used to calculate histogram of image gradients around key points to characterize the image feature. \par
To identify the interest point from the word picture in this study, we employed the SIFT technique. The size of the interest point is dynamically determined depending on the octave and scale data collected during interest point localization.
 
\section{CNN BASED FEATURE LEARNING}\label{sec5}
Once the writing patches have been identified by SIFT algorithm, they need to be mapped into features for  subsequent classification/identification task. Neural network based techniques in recent years have gained popularity as a go to method for feature learning in a variety of applications(eg Image classification, object detection\cite{Ren:2015}, image segmentation\cite{He:2017}), due to their ability to extract data-adaptive information from the input. A convolution neural network is a class of neural network which uses a set of convolution block containing a group of filters kernel to extract features from the preceding input block. The weights of a filter kernel are tuned continuously during training phase using back propagation to optimize the learned features.The convolution operation of a CNN layer can be mathematically expressed as:  
\begin{equation}
    x_{j}^{l}=f(\sum_{i\in M_{j}}x_{i}^{l-1} \times k_{ij}^{l}+b_{j}^{l})
\end{equation}
Here, $x_{j}^{l}$ represents the $j^{th}$ filter output map of layer $l$, $f$ represents the non-linear activation function, $M_j$ represents a section of the input feature map, $k$ represents the convolution filter kernel and $b$ represents the bias term. \par

Most of the existing CNN based writer identification algorithm uses image patches extracted from enrolled writer sample in a writer database to train the network \cite{semma:2021,He:2020,Javidi:2020,Semma:2022}. Such a writer dependent CNN network may not always be a viable option mainly because of the following reasons:
\begin{itemize}
    \item Due to the presence of multiple kernel in various convolution layers of CNN, these network require a large number of training sample to effectively discriminate various features of a writer. Since, the number of enrolled samples per writer in not uniform in a database, this may lead to high variation in the network performance across different writers \cite{Luo:2018}.
    \item Such a network is not effective in generalizing writer features effectively. Its performance is found to degrade when the writer samples used for testing are taken from a different database, as opposed to it being taken form the same database on which it was trained\cite{He:2020}. 
\end{itemize}

In an attempt to address the above mention shortcomings associated with a classification based convolution network, we use the concept of One shot learning \cite{Fei:2006,Lake:2011,Koch:2015} which changes a classification problem into a difference evaluation problem using Siamese neural network. Siamese networks aid in determining an unknown sample's identity based on measuring similarity/dissimilarity between the sample and enrolled samples. Siamese networks offers the following flexibility over a classical convolution network: (a) Fewer data is required to train the network making it robust to class imbalance problem, (b) capable of learning  visual attributes that aren't exclusive to a certain image, and (c) present a competitive strategy that is  not reliant on domain-specific expertise to learn the features form input.\par

\begin{figure}[ht] 
 \centering
\includegraphics[width=0.45\textwidth]{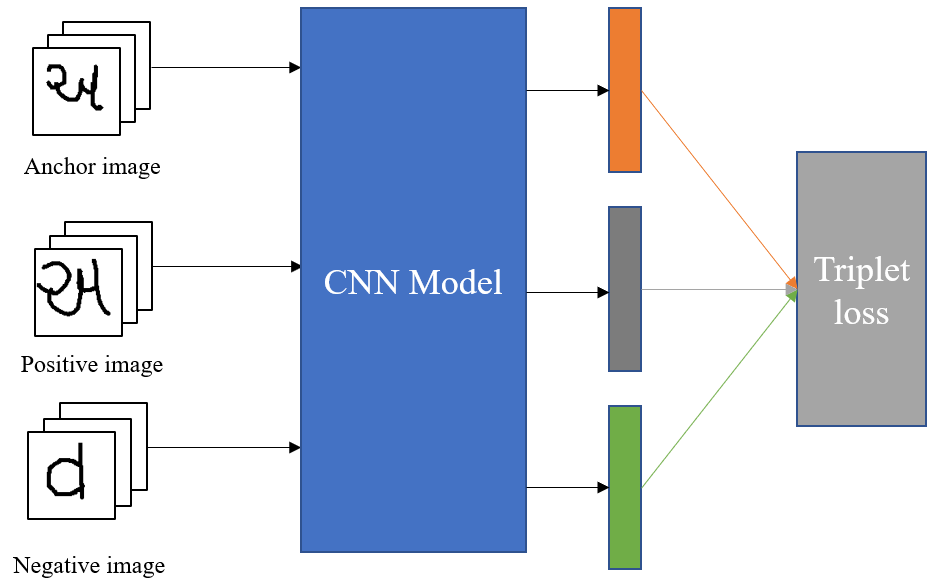}
\caption{Siamese network with triplet loss trained on Omniglot dataset \cite{Lake:2015}.}
 \label{fig:siamese_ntwk}
\end{figure}

\begin{figure}[ht] 
 \centering
\includegraphics[width=.48\textwidth]{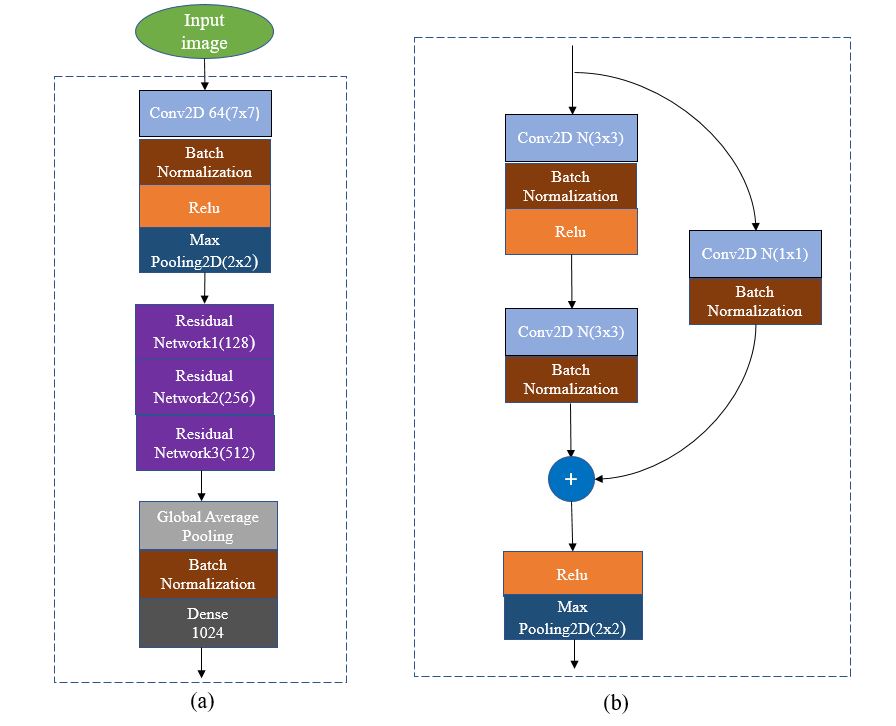}
\caption{An residual block based convolution Network used for extracting writer features. In this figure (a) represents the overall structure of convolution network, and (b) represent the architecture of the residual block used in the convolution network (Here, N represents the number of filter in the residual block).}
 \label{fig:sm}
\end{figure}

Siamese network consists of one or more identical network that receive a set of input and are connected at the top by an energy function, as shown in figure \ref{fig:siamese_ntwk}. This energy function is used to updates the network parameter during training. Weight sharing helps in ensuring that similar images are mapped to similar location in a feature space, thus increasing the classification ability of the network. The architecture of the convolution block used in our proposed Siamese network is shown in figure \ref{fig:sm}. Our networks uses the residual block (form Resnet-34 \cite{He:2016} architecture) in order to learn the input features. This network is trained using writer independent Omniglot dataset \cite{Lake:2015} containing 1623 different handwritten characters from 50 different alphabets, in order to capture diverse writing features across writers. Training a network on a writer independent database helps in better characterization of handwriting styles across different writer databases. This reduces the possibility of the convolution network remembering databases specific writer features. The weight of the Siamese network is updated using a category of loss function known as ranking loss. This category of loss function helps in modifying the weights of the convolution network, such that similar data are separated by a small distance compared to dissimilar data. For adjusting the weights of the Siamese network we have used a category of ranking loss known as triplet loss  \cite{Hoffer:2015}, consisting of two matching samples from the same class and a non-matching sample from a different class separated by a distance margin. Mathematically represented as:
\begin{equation}
\resizebox{.9\hsize}{!}{$\mathcal{L}\left(A,P,N\right)=\operatorname {max}\left(\|f(A)-f(P)\|^{2}-\|f(A)-f(N)\|^{2}+\alpha,0\right)$}
\end{equation}
Where $A$ is the anchor input, $P$ is the positive sample from the same class as that of $A$, $N$ is a negative input from class different from $A$, $\alpha$ is the threshold margin between the positive an negative pairs, and $f$ is the embedding function.\par

The images patches extracted at different scales from word image of a writer using SIFT algorithm is resized to a fixed size (105$\times$105) while maintaining the aspect ratio and padding it with white pixel until it reaches the predefined size. These modified image patches is then passed trough a Siamese network trained on Omniglot dataset. The features extracted from the outermost layer of one of the trained CNN model is used for feature extraction.
 
\section{FEATURE ENCODING}\label{sec6}
Most of the neural network based algorithm make their prediction based on multiple features. These features are centered on projecting input data into large dimensional space to increase separability among the input samples. Such a large number of features requires a lot of computational resource to process, in addition to that many of the features are often co-related or redundant which do not contribute to any useful information. Feature selection is used to mitigate this issue which focuses on reducing the number of random features under consideration. One of the most popular algorithm used to achieve this objective is Principle Component Analysis(PCA), which emphasises on representing input data in a set of principle components along which maximum variation in data is captured, without loosing useful information. However, one of the major limitation involving PCA  is related to its principle components being a linear combination of the input variables. This causes its corresponding projection coefficients to be non-zero while projecting the original data into lower dimensional space. As such it leads to difficulty in interpreting the relative importance of the derived principle components. \par

To mitigate this issue a need was felt to improve the interpretability of PCA by modifying its formulation such that along with reducing the dimensionality it also reduces the number of variables used to represent the original input. Such a representation leads to better interpretation of the input data, especially in cases where the number of input variable is large. Thus a variation of PCA called sparse PCA was formulated. Sparse PCA uses a variety of methods to introduce sparsity \cite{guerra:2021} in original PCA formulation. In our feature encoding block we use variant of Sparse PCA  which is formulated as a regression type problem and uses elastic net penalty to achieve sparse loading\cite{Zou04}. This variant of Sparse PCA is  mathematically represented as:
\begin{equation}
    \hat{\beta}=\underset{\beta}{\textit{arg\:min}}||Z_{i}-\textbf{X}\beta||^{2}+\lambda||\beta||^{2}+\lambda_{1}||\beta||_{1}
\end{equation}
Here, $Z_i$ is the target vector representing the principle component of each observation obtained by applying SVD on input matrix $X=UDV^{T}$ and rewriting it as $Z=UD=XV$. $\beta_{i}$ is the optimum solution of the above problem and  $\hat{\textit{V}_{i}}=\frac{\hat{\beta}}{||\hat{\beta}||}$ is the sparse approximation of the original loading vector $V_{i}$. These loading vector are combined to for a set of  sparse  PCA vector having $L$ components.
\begin{equation}
    \hat{V}=[\hat{V}_1 \;\hat{V}_2\; ... \;\hat{V}_l\; ... \;\hat{V}_L]
\end{equation}
The output from one of the branches of the Siamese network is project using these set of approximated sparse principle component  as follows:
\begin{equation}\label{projection}
    \alpha=X\hat{V}
\end{equation}
 In the matrix form a can be written as:
 \begin{equation}\label{projection_coeff}
\alpha = 
\begin{pmatrix}
\alpha_{1,1}^{i} & \alpha_{1,2}^{i} & \cdots & \alpha_{1,L}^{i} \\
\alpha_{2,1}^{i} & \alpha_{2,2}^{i} & \cdots & \alpha_{2,L}^{i} \\
\vdots  & \vdots  & \vdots & \vdots  \\
\alpha_{N,1}^{i} & \alpha_{N,2}^{i} & \cdots & \alpha_{N,L}^{i} 
\end{pmatrix}
\end{equation}

Here, $\alpha_{j,k}^{i}$ represents the multiplicative factor corresponding to the $k^{th}$ Sparse component for the $j^{th}$ segmented fragment sample of $i^{th}$ writer. Here, $N$ represents the number of fragments extracted from an individual word image.\par

\section{Assigning significance score for sparse components}\label{sec7}
In this section we discuss in details the methodology employed to assign a significance score to each spares components of modified PCA. Our methodology draws inspiration from the work carried out by Venugopal \textit{et.al} \cite{venugopal2018}. Let $N_{T_r}$ denote the total number of word fragments extracted from $W$ writers, with each writer contributing $\{n_{i}\}_{i=1}^W $ number of word fragments such that $\sum_{i=1}^{W}n_{i}=N_{T_r}$. \par

Based on the $N_{T_r}$ word fragments we carry out the following set of operations to assign a significance score to each sparse component:
\begin{enumerate}
    \item Constructing a histogram for each component of approximated sparse PCA.
    \item Computing relative-entropy value using the histogram generated in above step.
    \item Using the normalized relative-entropy value to assign a significance score to each sparse component.
\end{enumerate}
In the subsequent subsection each of the above mentioned step is discussed in detail.

\subsection{Histogram Generation}
 The first step in the process of histogram generation includes projecting $N_{T_r}$ word fragments obtained form $W$ writer samples onto the set of Sparse PCA components using equation \ref{projection}. As a result of this projection we obtain a set of sparse coefficient corresponding to each of the $N_{T_r}$ fragments as in equation \ref{projection_coeff}. Each column of this matrix signifies the set of $N_{T_r}$ multiplicative coefficients corresponding to a particular component of the sparse vector.\par 

 Sparse vector coefficient generated by $n_i$ fragments contributed by a specific writer is used to construct a histogram using the following equation:
\begin{align}\label{hist}
    H^{i}_{kb}=
\begin{cases}
    H^{i}_{kb}+1,& \text{if } h_{b}\leq\alpha_{j,k}^{i}<h_{b+1} \\
    H^{i}_{kb},              & \text{otherwise} 
\end{cases}\\
   1\leq j\leq n_i,\; 1\leq b\leq B \nonumber
\end{align}
Here, $H^{i}_{kb}$ signifies the sparse coefficient $\alpha_{j,k}^{i}$ representing the $j^{th}$ fragment of writer $i$ being assigned to $b^{th}$ quantized bin. $k$ represents the component of the sparse vector, and $B$ represents the total number of quantized bins. The size of each is determined based on Freedman–Diaconis rule \cite{Freedman1981}.

As a result of applying above operation we obtain a set of $W$ histogram associated with each component of sparse vector. These histogram are then normalized in the range between 0 to 1 by applying the following operation:
\begin{equation}
    p^{i}_{kb}=\frac{H^{i}_{kb}}{\sum^{B}_{b=1}H^{i}_{kb}}
\end{equation}

\subsection{Assigning significance score based on relative-entropy}
The normalized histogram $p^{i}_{kb}$ is uses to compute relative-entropy between $W$ writer samples for each sparse component using Kullback–Leibler (KL) divergence score \cite{kullback1951}. KL divergence score measures how much a probability distribution varies from another. We compute the KL divergence score between normalized histogram distribution as:
\begin{align}
    D_{j,k}^{i}=\sum_{b=1}^{B}p^{i}_{kb}\:\mathrm{log_2}\:\bigg(\frac{p^{i}_{kb}}{p^{j}_{kb}}\bigg)
\end{align}
Here, $D_{j,k}^{i}$ represents the divergence score between histogram distribution of writer $i$ and $j$ for $k^{th}$ sparse component. As a result of applying this divergence operation we get a set of divergence score, which can be represented using the following divergence matrix:
\begin{equation}\label{Entr}
D_k = 
\begin{pmatrix}
D_{1,k}^{1} & D_{2,k}^{1} & \cdots &D_{W,k}^{1} \\
D_{1,k}^{2} & D_{2,k}^{2} & \cdots &D_{W,k}^{2} \\
\vdots  & \vdots  & \ddots & \vdots  \\
D_{1,k}^{W} & D_{2,k}^{W} & \cdots &D_{W,k}^{W}
\end{pmatrix}
\end{equation}

Each entry along the rows of this divergence matrix represents degree of dissimilarity between a writer histogram distribution with respect to all other enrolled writers. Adding all the values of this divergence matrix gives the average divergence score for the $k^{th}$ sparse component represented as
\begin{equation}
    \phi_k=\frac{\sum_{i=1}^W \sum_{j\neq i}^{W}D^{i}_{j,k}}{W*(W-1)}
\end{equation}

The average divergence value of all the other sparse component can be calculated in a similar way. The set of obtained divergence score is used to assign a significance score to each sparse component through an inverse relationship specified as:
\begin{equation}
    w_k=\frac{1}{1+\phi_k}
\end{equation}

The motivation of assigning a significance score to a sparse component which is inversely related to the divergence value can be explained by the proposition, that a high divergence value (for a spares component ($\phi_k$)) signifies a high amount of relative uniformity among the samples. This, makes the classification task difficult, thus resulting in it being assigned a lower significance score.
Contrary to this a low divergence score signifies that there exist some degree of diverseness among samples which aides classification, resulting in it being assigned comparatively higher significance score.\par

\begin{figure*}[ht] 
\centering
\includegraphics[width=\textwidth]{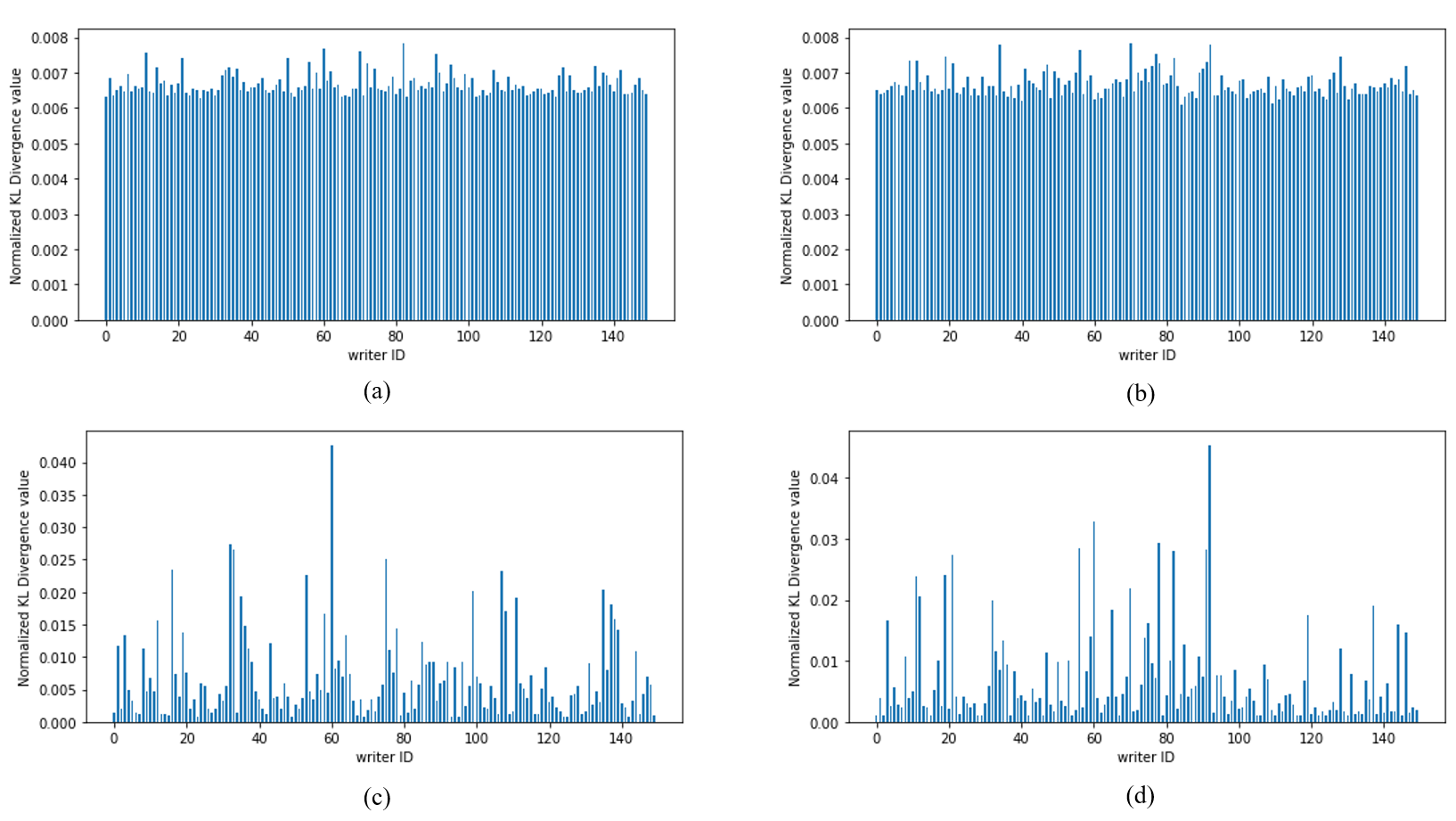}
\caption{(a) and (c) represents histogram corresponding to sparse component having maximum and minimum divergence value. (b) and (d) represents the histogram corresponding to the second maximum and minimum divergence value.}
 \label{saliency}
\end{figure*}

In order to support our observation, we present a visual representation of normalised divergence score corresponding to four sparse component in Fig. \ref{saliency}. For constructing the divergence histogram we have utilized SIFT fragments corresponding to 150 writers from IAM \cite{IAM}
 writer database. \par
 
\section{Generating Modified Writer Descriptor and classification}\label{sec8}
 Given a handwritten document containing $N_w$ word samples, generating $n_T$ word fragments using SIFT algorithm. Output of these fragments upon passing through a Simaese based CNN network after being projected onto a set of Sparse PCA components can be represented with the help of a matrix of size $(n_T \times K)$. The $(j,k)$ entry of this matrix represents the coefficient of the $j^{th}$ word fragment relative to the $k^{th}$ sparse component. These spare component are multiplied with the significance score assigned to each sparse component, thus generating a modified feature representation for each of the $n_T$ word fragments.   
 \begin{equation}
     \hat{Z}=Z\circ \hat{\phi}
 \end{equation}
 Here, $\hat{Z}$ represents the entries of the modified input matrix containing the sparse representation of the $n_T$ word fragments obtained by carrying element wise product of the original matrix $Z$ with weight matrix $\hat{\phi}$.\par
 
 The generated modified sparse representation is used to construct a one vs all SVM classifier using radial basis function (RBF) kernel \cite{Burges:1998} to train the writer samples. Grid search is to find the optimal value of RBF parameters $C$ and $\gamma$.\par
 
 In the testing phase given a segmented word image containing $N_w$ word fragment. A modified descriptor is generated for each fragment, which are then passed through a set of SVM classifiers trained individually for each enrolled writer in the database. These classifiers depending on the proximity of the testing samples to the separating hyperplane assign a score to each input fragments. Set of scores generated by each classifier is normalized between $[0-1]$ by passing them through a sigmoid function. Finally the normalized classification score generated for each of the fragment is combined to obtained overall classification score relative to a set of SVM classifiers.
 \begin{equation}
     P_l(word)=\frac{1}{N_w}\sum_{i=1}^{N_w}p_{f_i}
 \end{equation}
 Here, $P_l(word)$ is the overall classification score generated for a word image when passed through SVM classifier trained on $l^{th}$ writer sample. $N_w$ is the number of fragments associated with the input word image, and $p_{f_i}$ is the classification score of each word fragment. \par
 
The final predictor score for an input word image based on the classifier label $(l)$ for which highest classification score is observed.
\begin{equation}
 \hat{y}=\underset{l\in \{1,...,W\}}{argmax}P_{l}(word)
\end{equation}

\section{Dataset description}\label{sec9}
The proposed method is evaluated on two datsets: IAM \cite{IAM} and CVL \cite{CVL}.\par
The IAM database is one of the most prominent database used in the field of writer identification/verification. This database contains handwriting samples collected from 657 writers. Each writer has contributed a variable number of handwritten documents, 301 writers have contributed more than two handwritten document while the rest of 301 writers have contributed only one handwritten document. In order to evenly accumulate the handwritten samples from different writers, we randomly select two document per writer for writers that have contributed two or more document and split the document approximately in half for writers who have contributed only one document. Half of the accumulated samples per writer is used for training while the other half is used for testing as done in \cite{Khan:2018}. \par

The CVL dataset contains handwritten document collected from 310 writers of which 27 writers have contributed 7 documents (6 in English and 1 in German) and the rest 284 have contributed 5 documents (4 in English and 1 in German). In our experiment we have utilized only handwritten English document. In order to divide the handwritten documents in training and testing sets, we have followed the methodology employed in \cite{Khan:2018} which used three document per writer for training and one for testing.\par
\begin{figure}[ht] 
 \centering
\includegraphics[width=.48\textwidth]{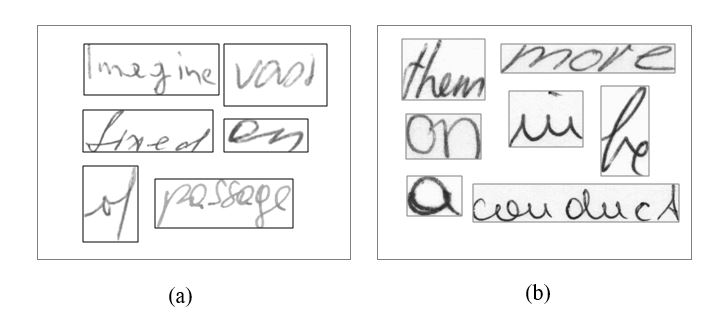}
\caption{Examples of training samples in (a) CVL database, and (b) IAM database. }
 \label{fig:db}
\end{figure}

    
Table \ref{datset} gives a detailed overview of the datasets used in our experiments and Fig.\ref{fig:db} shows some of the training sample of the dataset. In both the dataset segmented word images is provided by the author of the respective datasets.
 
\begin{table}[t]
\centering
\caption{Overview of the datasets used in experiments}
\label{datset}
\begin{tabular}{@{}ccc@{}}
\toprule
Dataset   & Number of writers & Language \\ \midrule
IAM       & 657               & English  \\
CVL       & 310               & English  \\ \bottomrule
\end{tabular}
\end{table}

\section{Experiment and Discussion}\label{sec10}
In this section we discuss in detail some of the experiments carried out by us on word image data for the above discussed database. These set of experiments helps us in choosing an optimum set of parameters, which helps in increasing the efficacy of our proposed writer identification system. 

\subsection{Implementation details}\label{CNN_impl}
\begin{figure*}[t] 
 \centering
\includegraphics[width=\textwidth]{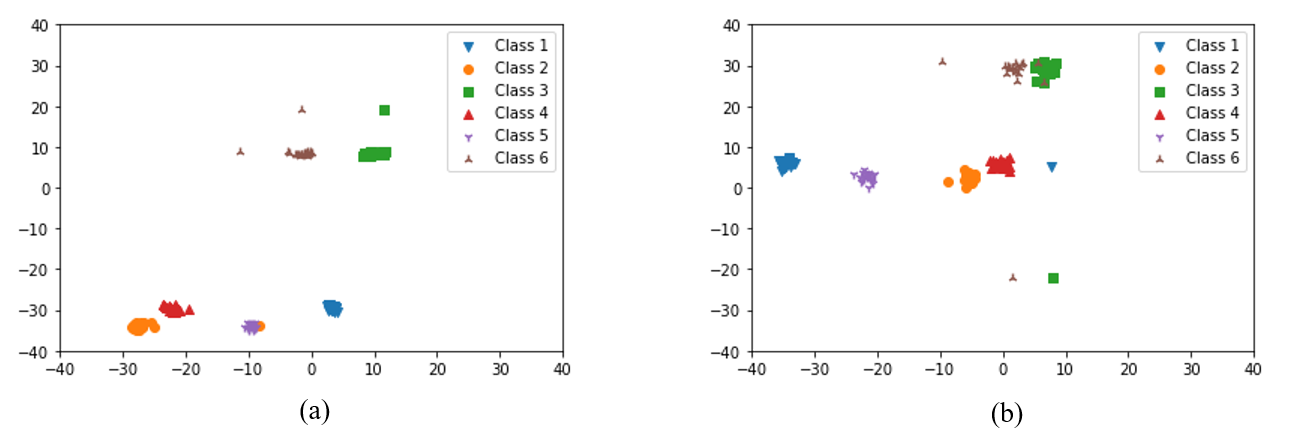}
\caption{2D projections of Omniglot dataset using t-SNE of feature vectors corresponding to, (a) Contrastive loss, and (b) Triplet loss. Input images are fed to the one of the Siamese block, which provides a D-dimensional feature vector representation. These feature vector are mapped to a 2-D representation using $t$-SNE projection methodology.}
 \label{fig:loss}
\end{figure*}

Our Siamese based Neural network is trained using Adam optimizer \cite{kingma:2014}. The batch size selected for training the network is set to 16. The learning rate is set to 0.001 per epoch. Selection of optimum rank loss function for training the weights of the network is based on the analysis two popularly used loss function namely contrastive \cite{hadsell:2006} and triplet loss. In our experimentation we have trained our network on Omniglot dataset using contrastive and Triplet loss. Fig.\ref{fig:loss} shows the effectiveness of contrastive and Triplet loss in classifying samples from the Omniglot dataset. The clustering efficiency of the loss function is measured by using Davies-Bouldin Index (DBI) \cite{DB:1979}, which measures the ratio of within-cluster distances to between-cluster distance. The DBI score for contrastive and triplet loss for a set of samples takes from Omniglot dataset is 1.12 and 1.27 respectively. A high DBI score being assigned to contrastive loss function as opposed to the triplet loss can be explained by the reasoning put forward by Schroff \textit{et.al} \cite{schroff:2015}, according to which contrastive loss forces all the samples of a positive class to converge to a single point in the embedding space which in some cases leads to a loss of intraclass information. On the other hand  triplet loss encourages the network to learn intraclass features by enforcing margin between a  pair positive class samples besides ensuring interclass separability among samples. Based on this explanation we select  Siamese network trained using triplet loss function for extracting writer features.  

\subsection{Sensitivity of Siamese network to projecting space Dimensions} 
In this experiment we analyze the effect of fully connected layers on the accuracy of the writer identification system. We vary the number of fully connected layers from 256 to 2048, doubling it at each step. Table \ref{inetr_result} lists the effect of varying the number of fully connected layer on the writer identification accuracy tested on IAM and IBM dataset respectively. In our analysis we have randomly selected a set of 100 writer from both dataset. The best average accuracy rate achieved is represented in bold font corresponding to each dataset. \par
From table\ref{inetr_result} we observe that the optimal number of dimension for classifying writer samples of IAM and CVL datset is 1024 and 512 respectively, as the number dimension of fully connected layer is increased the performance of the system deteriorates. This variation in the number of fully connected layer to represent the writer samples can be explained based on the following observation made with respect to the respective datasets. In CVL dataset the number of samples contributed by the enrolled writers are distributed uniformly across writers, opposed to this in case of IAM the number of samples contributed by each writer varies across the dataset. Due to this the impact on writing style is felt more in the case of IAM as opposed to CVL resulting in variable number of fully connected layers for these datasets.

\begin{table}[ht]
\centering
\begin{tabular}{|c|cccc|}
\hline
\multirow{2}{*}{Database} & \multicolumn{4}{c|}{Number of  FC layer}                                                     \\ \cline{2-5} 
                          & \multicolumn{1}{c|}{256}   & \multicolumn{1}{c|}{512}   & \multicolumn{1}{c|}{1024}  & 2048  \\ \hline
IAM                       & \multicolumn{1}{c|}{90.69} & \multicolumn{1}{c|}{91.69} & \multicolumn{1}{c|}{\textbf{91.89}} & 90.89 \\
CVL                       & \multicolumn{1}{c|}{81.41} & \multicolumn{1}{c|}{\textbf{84.44}} & \multicolumn{1}{c|}{84.04} & -     \\ \hline
\end{tabular}
\caption{Table displaying the effect of number of fully connected layer on writer performance.}
\label{inetr_result}
\end{table}

\begin{table}[ht]
\caption{Comparison of average writer identification rate (in \%) on word level data}
\label{compr}
\begin{tabular}{c|cc|cc}
\hline
\multirow{2}{*}{Methodology employed} & \multicolumn{2}{c|}{IAM}           & \multicolumn{2}{c}{CVL}            \\
                                      & \multicolumn{1}{c|}{Top1}  & Top5  & \multicolumn{1}{c|}{Top1}  & Top5  \\ \hline
Traditional method                    & \multicolumn{1}{c|}{85.22} & 92.39 & \multicolumn{1}{c|}{74.5}  & 89.32 \\
sparse based representation      & \multicolumn{1}{c|}{86.55} & 93.10 & \multicolumn{1}{c|}{77.34} & 91.06 \\
Incorporating sparse weighting        & \multicolumn{1}{c|}{86.78} & 93.15 & \multicolumn{1}{c|}{78.00} & 91.39 \\ \hline
\end{tabular}
\end{table}

\subsection{Influence of Sparse and weighted sparse}
In this experiment we study the effect of incorporating the significance score to the sparse based representation and its effect on overall accuracy of the system. As a part of the experiment we compare the performance of the sparse and modified sparse based representation over the traditional method (consisting of training an SVM classifier on the output of Siamese network). The result obtained on word level data is shown in Table \ref{compr} for IAM and CVL databases.\par
 From table \ref{compr} we observe that as a result of incorporating significance based score into the sparse representation the performance of the system increases. For IAM database the accuracy increases from 85.22\% using traditional method to 86.78 \% by incorporating significance score when tested on word level data. Similarly for CVL dataset the accuracy increases form 74.5\% to 78.00\%. This result strengthens our observation made in accordance with figure \ref{saliency}.\par
 
\begin{figure*}[ht] 
 \centering
\includegraphics[width=\textwidth]{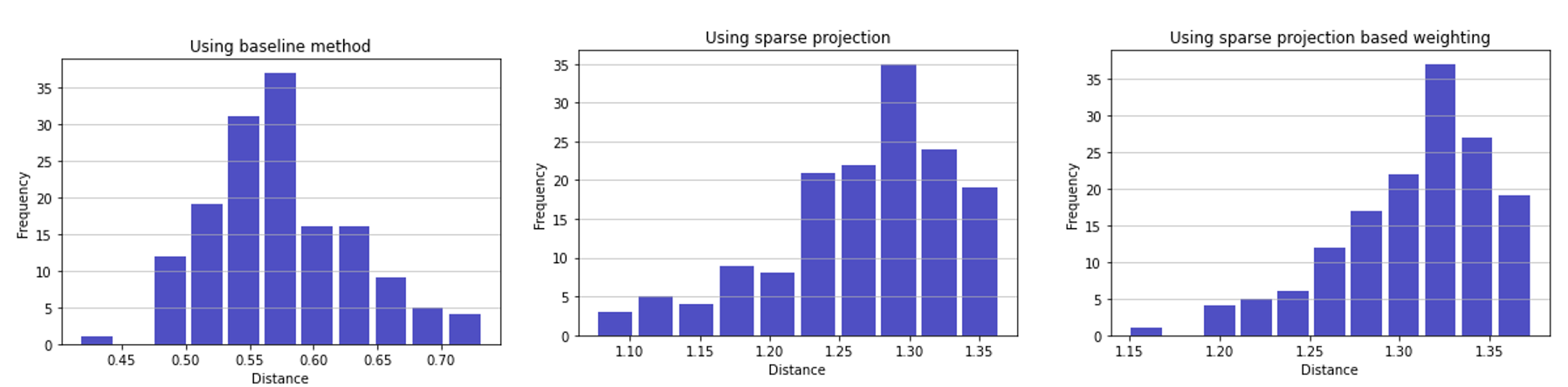}
\caption{ Average Intra-class distance between writer samples}
 \label{intra_class}
\end{figure*}

 As an additional support demonstrating the impact of sparse PCA on the feature representation of a writer fragment, we extract the features relative to a group of writer fragments collected from a subset of writers by passing these fragments through the trained convolution network containing the penultimate layer. These, features are then modified using the above discussed sparse and modified sparse representation, to generate a histogram based representation signifying the average intra-class distance for a set of fragments related to each writer. Fig. \ref{intra_class} shows the result of this analysis done on word samples collected from IAM databse. Based on fig.\ref{intra_class} it can be observed that average intra-class distance is minimum for baseline method signifying that the samples are densely packed in an embedding space. This may leads to loss of intra-class information resulting in a lower accuracy score when compared to Sparse based projection, which increases the separability among the data samples leading to a higher intra-class distance. 
 
\begin{figure*}[ht] 
 \centering
\includegraphics[width=\textwidth]{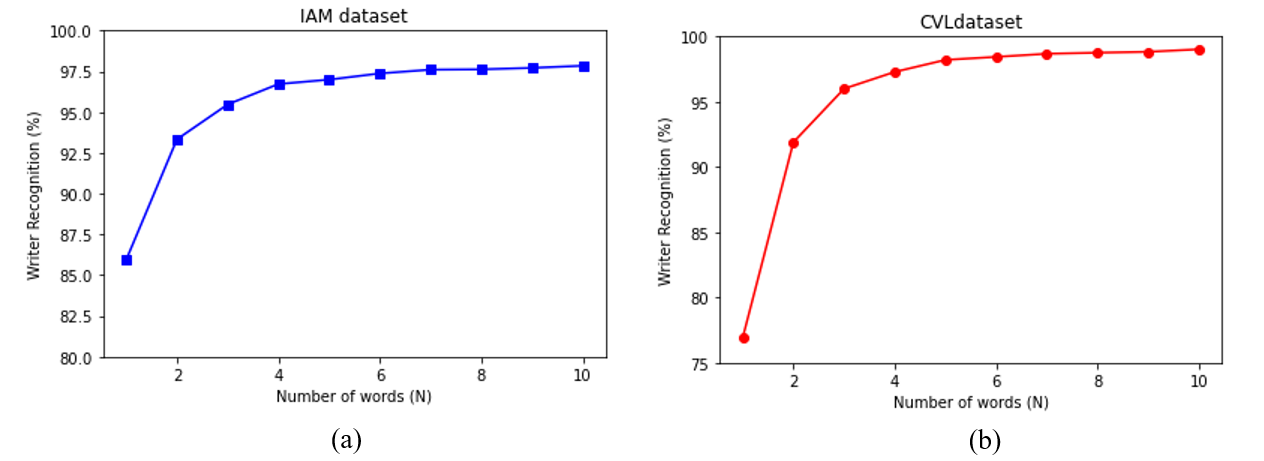}
\caption{Performance of proposed writer identification system with regard to availability of varying number of word image tested on, (a) IAM, and (b) CVL dataset respectively.}
 \label{var_word}
\end{figure*}

\subsection{Performance of Writer Identification with varying number of word image}
In this experiment we show the performance of our writer identification system with respect to the amount of handwritten data available for classifying writer samples. The objective of this analysis is to show the effectiveness of our proposed system in scenarios where limited amount of writer data is available for identifying a writer sample (as shown through fig.\ref{fig:limted word}). Fig. \ref{var_word} shows the performance of our writer identification system when tested on IAM and CVL dataset with different amount of word image. \par
It can inferred form Fig.\ref{var_word} that as the available number of word per writer increases from one to two the overall performance of the system shows a steep rise. This improvement in performance is attributed to the overall increases in the allographic information availability, thus increasing the classification accuracy of the system. This increase in accuracy attains a saturation level corresponding to four words per writer, which signifies the minimum number of words required by our algorithm to identify a writer with a high confidence score.

\subsection{Performance comparison of proposed system with handcrafted and deep learning features}\label{subsec:handcrafted}
In this section we evaluate the performance of our system with other traditional handcrafted feature as well as deep-learning based algorithm trained on a word image data. Table \ref{result1} summarizes the performance of various algorithm considering segmented word image as input. Based on table \ref{result1} it can be observed, that in comparison to handcrafted feature based algorithm deep neural network achieves higher accuracy when trained on word images. This disparity in performance can be attributable to the fact that the majority of conventional writer identification algorithms utilise handcrafted features to extract relevant data from the input image based on performing statistical analysis of the features. As, such for generating a stable representation of a writer data, these method require a certain minimum amount of text samples. In the case of word image the extracted handcrafted feature  are insufficient to obtain a stable representation of a writer sample. Contrary to this the presence of multiple convolution layer along with the concept of weight sharing enable a convolution based neural network to learn diverse and distinctive features from a word image resulting in better performance over traditional methods. \par

 From table\ref{result1}, it can be inferred that our system outperforms other deep learning techniques trained on word image for the IAM database compared to CVL database. The following findings on both databases provide an explanation for this performance disparity.:
\begin{itemize}
    \item  In IAM dataset each writer contributes variable amount of text image, with more than half of the writer (350)  contributing only one page, on the other hand in CVL dataset the writer samples are almost equally distributed among enrolled writers.
    \item In CVL dataset the content of text is same across different writer.
\end{itemize}
As a result the effect of writing style is more prominent in IAM dataset compared to CVL datset. Moreover our CNN network in trained on writer independent dataset, it is sometimes unable to differentiate confidently between writer having similar writing style (due to the content being the same across the writers) leading to false acceptance.

\begin{table*}[ht]
\centering
\caption{Comparison of Writer Identification performance (in \% ) on full database for word image samples}
\label{result1}
 %\resizebox{\textwidth}{!}{%
\begin{tabular}{ccccccc}
\hline
\multirow{2}{*}{Method}                  & \multicolumn{2}{c}{IAM}                     & \multicolumn{2}{c}{CVL}                                 \\ \cline{2-5} 
                                         & Top1                 & Top5                                  & Top1                 & Top5                                \\ \hline
Hinge\cite{Bulacu:2007}                                    & 13.8                 & 28.3                 & 13.6                 & 29.7                                                  \\
Quill\cite{Brink:2012}                                    & 23.8                 & 44.0                 & 23.8                 & 46.7                                                  \\
CoHinge\cite{Sheng:2017}                                 & 19.4                 & 34.1                 & 18.2                 & 34.2                                                  \\
QuadHinge\cite{Sheng:2017}                                & 20.9                 & 37.4                 & 17.8                 & 35.5                                                   \\
COLD\cite{sheng:2017C}                                    & 12.3                 & 28.3                 & 12.4                 & 29.0                                                  \\
Chain Code Pairs\cite{Siddiqi:2010}                         & 12.4                 & 27.1                 & 13.5                 & 30.3                                                   \\
Chain Code Triplets\cite{Siddiqi:2010}                       & 16.9                 & 33.0                 & 17.2                 & 35.4                                                   \\
WordImgNet\cite{He:2020}                                & 52.4                 & 70.9                 & 62.5                 & 82.0                                                   \\
FragNet-64\cite{He:2020}                               & 72.2                 & 88.0                 & 79.2                 & 93.3                                                  \\
Veritcal GR-RNN(FGRR)\cite{He:2021}                               & 83.3                 & 94.0                 & 83.5                 & 94.6                                                  \\
Horizontal GR-RNN(FGRR)\cite{He:2021}                                 & 82.4                 & 93.8                 &82.9                 & 94.6                                                  \\\hline
Proposed Methodology (sparse representation + divergence score)  & 86.78 & 93.15 & 78.0 & 91.39  \\\hline

\end{tabular}%}
\end{table*}

\subsection{Performance comparison of proposed system on page Image}

\begin{table*}[ht]
\centering
\caption{ Comparison of State of art methods on IAM database}
\label{sota_IAM}
 %\resizebox{\textwidth}{!}{%
\begin{tabular}{ccc}
\hline
Reference                                  & Number of Writers & Top 1 accuracy \\ \hline
Siddiqui and Vincent\cite{Siddiqi:2010}                      & 650               & 91.0           \\
He and Schomaker\cite{Sheng:2017}                           & 650               & 93.2           \\
Khalifa et.al.\cite{Khalifa:2015}                             & 650               & 92.0           \\
Hadjadji and Chibani\cite{Hadjadji:2018}                       & 657               & 94.5           \\
Wu et.al.\cite{Wu:2014}                                 & 657               & 98.5           \\
Khan et.al.\cite{Khan:2018}                               & 650               & 97.8           \\
Nguyen et.al.\cite{Nguyen:2018}                              & 650               & 93.1           \\
WordImagenet                               & 657               & 95.8           \\
FragNet-64\cite{He:2020}                                & 657               & 96.3           \\
GR-RNN\cite{He:2021}                                   & 657               & 96.4           \\
Proposed Methodology (sparse representation + divergence score)   & 657               & 97.84               \\ \hline
\end{tabular}%}
\end{table*}

\begin{table*}[ht]
\centering
\caption{Comparison of state of art method on CVL database}
\label{sota_CVL}
 %\resizebox{\textwidth}{!}{%
\begin{tabular}{ccc}
\hline
Reference                                  & Number of Writers & Top 1 accuracy \\ \hline
Fiel and sablating\cite{Fiel:2015}                        & 309               & 97.8           \\
Tang and Wu\cite{Tang:2016}                               & 310               & 99.7           \\
Christlein et.al\cite{Christlein:2017}                           & 310               & 99.2           \\
Khan et.al.\cite{Khan:2018}                                & 310               & 99.0           \\
WordImagenet                               & 310               & 98.8           \\
FragNet-64\cite{He:2020}                                 & 310               & 99.1           \\
GR-RNN\cite{He:2021}                                     & 310               & 99.3           \\
Proposed Methodology (sparse representation + divergence score)   & 310               & 99.03              \\ \hline
\end{tabular}%}
\end{table*}

Table \ref{sota_IAM} and \ref{sota_CVL} tabulates the performance comparison of the proposed system with other writer identification algorithm based on page images. The overall writer accuracy is calculated by averaging the response of all the segmented word image obtained from a page image of the writer sample. This process can be mathematically represented as:
\begin{equation}
    P_{page}=\frac{1}{N_{page}}\sum_{t\in page}^{N_{page}}P(t)
\end{equation}
Here, $P_{page}$ represents the overall writer classification score generated corresponding to a trained writer classifier. $P(t)$ is the individual classifications score relative to each word $t$ contained in the page and $N_{page}$ is the total number of words in the page image.  \par

On the basis of Table \ref{sota_IAM} and \ref{sota_CVL} it can be observed that due to the availability of large number of word samples the performance of the proposed system is higher for page image as compared to word image. In case of CVL database no significant deviation in performance is observed when compared against major writer identification algorithm for page level data. However, for IAM database a significant improvement in performance is observed when compared against other deep learning based method (such as WordImgNet\cite{He:2020}, FragNet-64\cite{He:2020}, GR-RNN\cite{He:2021} and Nguyen et.al.\cite{Nguyen:2018}) tested on page images. This increase in accuracy can be attributed to the reasoning put forward in subsection \ref{subsec:handcrafted} with respect to CVL and IAM databases respectively. However, directly comparing these methodologies may not be considered fair as these algorithm employ diverse set of features, classifiers along with training and testing methodologies, which may affect their overall performance.

\section{Conclusion}\label{sec11}
Through this paper, we have tried to explore the effectiveness of a writer independent model in classifying a writer based on word image. In this work, we have used a Siamese network trained on writer independent omniglot datset \cite{Lake:2015} to extract the writer features from input samples. The performance of such type of network trained using writer independent approach is found to be at par to a network trained directly on writer data. Besides such a network needs minimum amount of retraining and can be used in scenarios where amount of writer data is limited as in fig \ref{fig:limted word}. In addition to this, we have utilized a sparse based variant of PCA called Sparse PCA to achieve sparsity in the fully connected layer of the Siamese network, each component which is assigning a significance score based on its divergence value. By incorporating these features the proposed idea is found to outperform the traditional baseline method, solely based on utilizing the output of fully connected layer of a neural network. This, methodology can be extended to on-line based writer and signature verification system trained using deep neural network framework. \par
The main objective of the proposed work is not focused to achieve state-of-the-art result, but to explore the possibility of using a network trained on writer independent data for the purpose of writer identification. Recent studies related to writer identification using neural network \cite{He:2021,He:2020} have used the fragments extracted from word level data along with the segmented word image as a whole to extract writer related information effectively. In our future work we would like to incorporate global feature across a writer fragments to enhance the overall accuracy of the writer identification algorithm.

%\appendices

%\section{}

%\section*{Acknowledgment}


\ifCLASSOPTIONcaptionsoff
  \newpage
\fi

\bibliographystyle{IEEEtran}
\bibliography{bibtex.bib}

\end{document}